\documentclass{article}

\usepackage{arxiv}
\usepackage{microtype}
\usepackage{booktabs}
\usepackage{amsmath}
\usepackage{amssymb}
\usepackage{amsthm}
\usepackage{mathtools}
\usepackage{enumitem}
\usepackage{hyperref}
\usepackage[capitalize,noabbrev]{cleveref}
\usepackage{natbib}
\usepackage{xcolor}
\usepackage{tikz}
\usetikzlibrary{arrows.meta,positioning,fit,calc}

\hypersetup{
  pdftitle={Agentic Skill Optimization over Lie Algebroids},
  pdfsubject={Infinitesimal causality, Lie algebroids, and agentic skill optimization},
  pdfauthor={Sridhar Mahadevan},
  pdfkeywords={infinitesimal causality, Lie algebroids, tangent categories, skill optimization, Markdown, backpropagation},
  colorlinks=true,
  linkcolor=blue!55!black,
  citecolor=blue!55!black,
  urlcolor=blue!55!black
}

\title{Agentic Skill Optimization over Lie Algebroids \thanks{Draft under revision.}}

\author{Sridhar Mahadevan\\
Adobe Research and University of Massachusetts, Amherst\\
\texttt{smahadev@adobe.com, mahadeva@umass.edu}}

\date{\today}

\newtheorem{definition}{Definition}
\newtheorem{proposition}{Proposition}
\newtheorem{conjecture}{Conjecture}

\newcommand{\M}{\mathcal{M}}
\newcommand{\Md}{\mathsf{Md}}
\newcommand{\A}{\mathcal{A}}

\newcommand{\D}{\mathcal{D}}
\newcommand{\V}{\mathcal{V}}

\newcommand{\Loss}{\mathcal{L}}
\newcommand{\IC}{\textsc{IC}}
\newcommand{\BRIDGE}{\textsc{BRIDGE}}
\newcommand{\SKFM}{\textsc{SKFM}}
\newcommand{\ALLORA}{\textsc{ALLORA}}
\newcommand{\LICKET}{\textsc{LICKET}}
\newcommand{\LASKO}{\textsc{LASKO}}
\newcommand{\SkillOpt}{\textsc{SkillOpt}}
\newcommand{\BASKET}{\textsc{BASKET}}
\newcommand{\ROCKET}{\textsc{ROCKET}}
\newcommand{\DB}{\textsc{DB}}
\newcommand{\doop}{\operatorname{do}}
\newcommand{\Curv}{\operatorname{Curv}}

\newcommand{\rank}{\operatorname{rank}}

\begin{document}
\maketitle

\begin{abstract}
Agentic systems increasingly improve themselves by editing skills: prompts,
rubrics, plans, tool contracts, examples, validators, and traces. Skill edits are not independent coordinates in a vector space: they are local repairs to structured artifacts whose effects are observed only after rollout, validation, and critique.  Distinct edits can have the same immediate visible effect while
differing in routing context, template state, guardrail scope, or future
composability.  The order of edits can matter as well: repairing a schema before
a normalization rule need not be equivalent to applying the same edits in the
reverse order.   This paper introduces a new framework for skill optimization called \LASKO, for Lie Algebroid SKill Optimization. \LASKO\ models
typed, anchored Markdown skills as the base category and available edit
policies as sections of a controlled Lie algebroid $\A\to\Md$ with anchor
$\rho$.  The anchor maps an edit policy to its visible Markdown effect; the
kernel $\ker(\rho)$ represents latent template, routing, or implementation
structure; and the algebroid bracket measures noncommuting edit composition.

As shown in the paper, \LASKO\ achieves order-of-magnitude speedups in skill optimization in our preliminary benchmark results, primarily because it substitutes inexpensive Lie-bracket screening tests that run in microseconds, before investing in expensive validations that require running large language models. On a causal extraction from natural language task, \LASKO\ achieved a speedup of almost $15 \times$ compared to a brute-force approach that validated all edits by running them through a DeepSeek V3.1 4-bit model with 671B parameters. 
\end{abstract}


\section{Introduction}

Agentic skill optimization, exemplified by the \SkillOpt\ framework of
\citet{yang2026skillopt}, is currently one of the most important frontiers of
agentic AI.  \SkillOpt\ makes a decisive step away from one-shot prompt writing
and loosely controlled self-revision: it treats a skill as external textual
state of a frozen agent, asks an optimizer model to propose bounded
add/delete/replace edits, and accepts an edit only when it improves held-out
validation performance.  This discipline has a practical virtue that is easy to
understate.  It turns agent improvement into an experimental loop: propose a
skill repair, roll out the agent, gate the edit on validation performance, and
repeat.

The difficulty is that the search space is not flat.  A skill is a structured
artifact: a prompt, rubric, schema, plan, tool contract, validator, example set,
or trace summary.  A failed rollout may be caused by any one of these anchors,
or by an interaction among several of them.  Two edits can look locally
reasonable and still fail as a pair; conversely, two weak edits may become
productive only in the right order.  Repairing a schema before a normalization
rule, adding an abstention policy before a formatting contract, or installing a
validator before a final-answer template are order-sensitive operations.  A
short-horizon \SkillOpt\ loop that validates only single edits can therefore
miss the actual repair, while exhaustive enumeration of ordered edit programs
quickly becomes the wrong place to spend expensive rollout calls.

This paper proposes that this is the central geometric problem of agentic skill
optimization.  We call the proposed generic framework \LASKO: Lie Algebroid
SKill Optimization.  The aim is not merely to formalize an already successful
method after the fact.  The aim is to expose the geometry that can improve
systems such as \SkillOpt: where an edit acts, which part of its effect is
visible, what hidden state it carries, how validation losses pull back to
artifact anchors, and which pairs of repairs fail to commute.  In this view,
the key object is not a scalar score for an isolated prompt edit, but the local
algebra of edits around a typed skill artifact.

Infinitesimal Causality (\IC)~\citep{mahadevan2026ic} supplies the first piece
of this algebra.  It replaces finite counterfactual guesses with local
interventions and studies their Lie brackets: if two infinitesimal
interventions fail to commute, their bracket records a local composition
defect.  Earlier work proposed Lie-algebraic methods for latent-confounded
causal discovery, specifically \BRIDGE\ (Bracket Residuals for Interventional
Discovery and Geometric Estimation) and \SKFM\ (Spectral Kan-Do Flow Matching),
which use bracket residuals to diagnose hidden causal structure
\citep{mahadevan2026bridge}.  Skill optimization needs the same idea, but with
one crucial lift: the controlled edit chosen by an optimizer is not identical to
the visible change observed in a document, rollout, model, or trace.

The remedy is to formulate \IC\ over Lie algebroids.  There is the
\emph{controlled intervention} chosen by an optimizer or agent, and there is the
\emph{visible tangent effect} observed on a structured skill artifact.  A
controlled intervention may have internal implementation coordinates, latent
state, routing context, or template variables that are not visible at first
order.  It may nevertheless bracket nontrivially with future interventions.
This suggests replacing the tangent bundle $T\M$ by a Lie algebroid
\[
  \A \longrightarrow \M,
  \qquad
  \rho : \A \longrightarrow T\M,
  \qquad
  [\,\cdot,\cdot\,]_\A :
  \Gamma(\A)\times\Gamma(\A)\to \Gamma(\A).
\]
The sections of $\A$ are controlled intervention modes.  The anchor $\rho$
maps a controlled intervention to its visible infinitesimal effect.  The kernel
$\ker(\rho)$ contains intervention structure with no immediate visible tangent
effect: latent, isotropy, or unanchored degrees of freedom.  The
algebroid bracket models the infinitesimal algebra of intervention composition
before this algebra is projected into observable dynamics.  Operationally, the
bracket becomes a screening statistic: high-bracket edit pairs are the pairs
worth validating when rollout calls are expensive.

For \LASKO, the base is not merely a Euclidean parameter space.  We model skills
in a tangent category of typed, anchored Markdown
objects: documents, rubrics, prompts, tool contracts, plans, examples,
validators, and traces.  A practical instance---for example a \SkillOpt,
\BASKET, \ROCKET, or Democritus workflow---edits these artifacts, rolls them
out, receives critiques or validation failures, and then edits again.  The
forward pass is the tangent functor: local edits push forward through rewrites,
rollouts, and observation contexts.  The backward pass requires reverse tangent
or cotangent structure: losses and critiques pull back to anchors.  A
controlled Lie algebroid over this Markdown base then captures which edit
policies are available, how they compose, and where hidden procedural structure
lives.  In a controlled ten-anchor benchmark we use this structure to
prioritize ordered edit pairs before validation.  The result is the empirical
wedge of the paper: a \LASKO-prioritized policy recovers the exhaustive
ordered-pair solution with a small validation budget, while greedy single-edit
search misses the interaction.

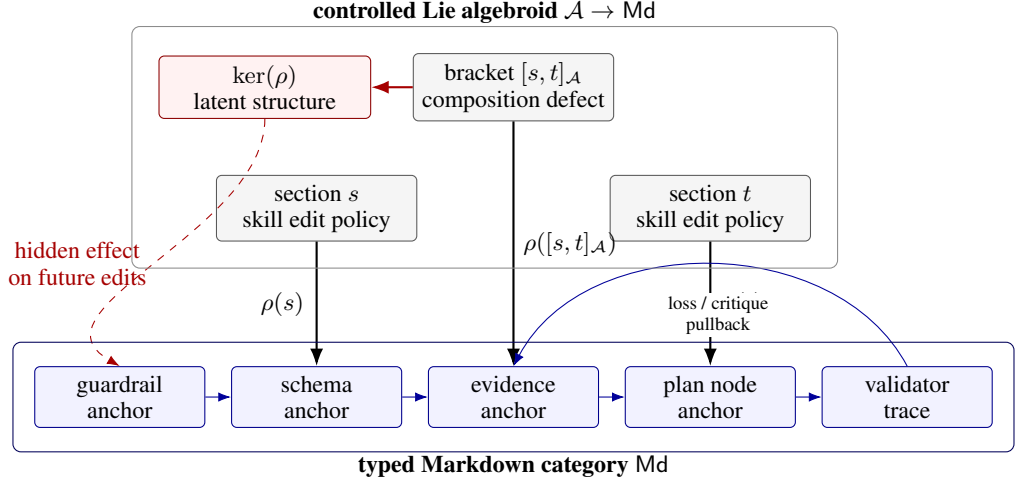
\begin{figure}[t]
  \centering
  \begin{tikzpicture}[
    font=\small,
    >=Latex,
    artifact/.style={
      draw=blue!55!black,
      fill=blue!5,
      rounded corners=2pt,
      minimum width=2.25cm,
      minimum height=0.72cm,
      align=center
    },
    sectionnode/.style={
      draw=black!65,
      fill=black!4,
      rounded corners=2pt,
      minimum width=2.65cm,
      minimum height=0.78cm,
      align=center
    },
    kernelnode/.style={
      draw=red!55!black,
      fill=red!5,
      rounded corners=2pt,
      minimum width=2.8cm,
      minimum height=0.78cm,
      align=center
    },
    boxlabel/.style={font=\small\bfseries, inner sep=1pt}
  ]
    \node[artifact] (guard) {guardrail\\anchor};
    \node[artifact, right=0.34cm of guard] (schema) {schema\\anchor};
    \node[artifact, right=0.34cm of schema] (evidence) {evidence\\anchor};
    \node[artifact, right=0.34cm of evidence] (plan) {plan node\\anchor};
    \node[artifact, right=0.34cm of plan] (trace) {validator\\trace};

    \node[sectionnode, above=1.65cm of schema] (s) {section $s$\\skill edit policy};
    \node[sectionnode, above=1.65cm of plan] (t) {section $t$\\skill edit policy};
    \node[sectionnode, above=1.15cm of $(s)!0.5!(t)$] (bracket)
      {bracket $[s,t]_{\A}$\\composition defect};
    \node[kernelnode, left=0.55cm of bracket] (kernel)
      {$\ker(\rho)$\\latent structure};

    \draw[->, thick] (s) -- node[left, xshift=-1pt] {$\rho(s)$} (schema);
    \draw[->, thick] (t) -- node[right, xshift=1pt] {$\rho(t)$} (plan);
    \draw[->, thick] (bracket) -- node[right] {$\rho([s,t]_{\A})$} (evidence);
    \draw[->, red!65!black, thick] (bracket) -- (kernel);
    \draw[->, dashed, red!65!black] (kernel.south) to[out=-90,in=150]
      node[left, align=center] {hidden effect\\on future edits} (guard.north);

    \draw[->, blue!60!black] (guard) -- (schema);
    \draw[->, blue!60!black] (schema) -- (evidence);
    \draw[->, blue!60!black] (evidence) -- (plan);
    \draw[->, blue!60!black] (plan) -- (trace);
    \node[font=\scriptsize, fill=white, inner sep=1pt, align=center]
      (pullbacklabel) at ($(evidence.north)!0.52!(trace.north)+(0,0.68cm)$)
      {loss / critique\\pullback};
    \draw[->, blue!60!black] (trace.north) to[out=118,in=62]
      (evidence.north);

    \node[draw=blue!35!black, rounded corners=3pt, fit=(guard)(schema)(evidence)(plan)(trace),
      inner xsep=8pt, inner ysep=9pt, label={[boxlabel]below:typed Markdown category $\Md$}]
      (basebox) {};
    \node[draw=black!45, rounded corners=3pt, fit=(s)(t)(bracket)(kernel),
      inner xsep=10pt, inner ysep=10pt, label={[boxlabel]above:controlled Lie algebroid $\A\to\Md$}]
      (algebroidbox) {};
  \end{tikzpicture}
  \caption{
    Agentic skill optimization as algebroidic infinitesimal causality over
    typed Markdown artifacts.  The base category consists of anchored
    documents, plans, tool contracts, and validator traces.  Sections of the
    controlled algebroid are available edit policies; the anchor $\rho$ maps an
    edit policy to its visible Markdown effect.  Brackets measure
    order-sensitivity of skill edits, while $\ker(\rho)$ records latent
    template, routing, or implementation structure that may not change the current
    artifact but can change future composition.
  }
  \label{fig:skillopt-algebroid}
\end{figure}

\Cref{fig:skillopt-algebroid} summarizes this specialization.  The lower row
is the typed Markdown base: anchors such as guardrails, schema fields, evidence
edges, plan nodes, and validator traces are the locations at which local
artifact changes and rollout observations are recorded.  The upper row is the
controlled algebroid of edit policies.  A section $s$ or $t$ is not itself a
visible Markdown change; its visible effect is obtained only after applying
the anchor $\rho$.  The bracket $[s,t]_\A$ records the infinitesimal
order-sensitivity of composing two skill edits, and its kernel component
represents latent template, routing, or implementation structure that may leave the
current document unchanged while changing how later edits behave.  The feedback
arrow from validator traces to evidence anchors depicts the reverse pass:
losses and critiques are transported back to the artifact locations where the
skill can be repaired.

\IC\ was originally formulated over ordinary tangent bundles.  The
hypothesis developed here is that, in structured applications such as agentic
skill optimization, the better formulation is over controlled Lie algebroids of
interventions.  \LASKO\ is the corresponding algebroid optimization problem
over a tangent category of agent-facing artifacts; \SkillOpt\ is one concrete
optimizer that can instantiate this geometry.

\section{Background}

Before developing the algebroid formalism, we fix the systems vocabulary and
the related lines of work.  \SkillOpt\ was introduced by
\citet{yang2026skillopt} as an executive strategy for self-evolving agent
skills.  We use the term in that spirit, and focus on the geometric structure
of the underlying optimization problem: an agentic system improves operational
skills by editing artifacts such as prompts, rubrics, plans, schemas, examples,
tool contracts, validators, traces, and critique summaries.  A \SkillOpt\
update is therefore not only a parameter update.  It is a structured
intervention on a typed artifact or workflow, followed by rollout, validation,
credit assignment, and another edit.  This paper asks for the analogue of
gradient geometry for that setting: what is the local direction of an edit,
where is its visible effect, what hidden state does it carry, and how do we
measure order-dependence among repairs?

The motivating agent-system examples are \BASKET\ and \ROCKET, introduced in
the broader diagrammatic agent-systems program of \emph{Categories for
AGI}~\citep{mahadevan2026catagi}.  For the purposes of this paper, \BASKET\
should be read as a workflow-construction mechanism: it turns source material
such as filings, evidence fragments, or task descriptions into a typed diagram
of claims, risks, plans, tool calls, and synthesized narratives.  \ROCKET\
should be read as a local workflow-repair mechanism: it searches near such a
diagram by repairing plan nodes, replacing weak evidence edges, adding missing
validators, or changing synthesis routes.  These systems motivate why skills
should be modeled as partially ordered diagrams of anchored artifacts rather
than as flat prompt strings.

The causal lineage comes from structural causal models and do-calculus
\citep{pearl2009causality}, potential-outcome causal inference
\citep{rubin2005causal}, and constraint-based causal discovery
\citep{spirtes2000causation}.  \IC~\citep{mahadevan2026ic} reframes
interventions infinitesimally: interventions become local vector fields, and
their Lie brackets measure local order-dependence.  \BRIDGE\ and
\SKFM~\citep{mahadevan2026bridge} use bracket residuals as diagnostics for
latent or missing causal directions.  In this paper, those ideas are imported
into \LASKO: a skill edit is treated as an intervention, validator failures
provide observable consequences, and noncommuting edits signal hidden structure
or missing anchors in the workflow.

The geometric and categorical lineage supplies the technical language for this
move.  Lie algebroids and their integrability theory
\citep{mackenzie2005general,crainic2003integrability} separate a controlled
intervention from its visible anchored effect by means of an anchor map
$\rho:\A\to T\M$.  Tangent categories provide the categorical foundation for
forward derivative structure~\citep{cockett2014differential}, while involution
algebroids~\citep{burke2019involution} provide a tangent-category analogue of
involutive closure.  In the present paper, the base object is usually not a
smooth manifold but a typed Markdown workflow.  The role of the algebroid is to
model the controlled edit policies over that workflow, their visible anchored
effects, and their hidden kernel structure.

Finally, neural-adapter composition is a useful parallel example, but not a
target of this paper.  Low-rank adaptation~\citep{hu2022lora} motivates
interventions on neural models, and \ALLORA/\LICKET~\citep{mahadevan2026allora}
study bracket-control objectives for composing such adapters.  Those ideas
support the general intuition that controlled interventions can carry hidden
structure above their visible effect.  Here, however, the worked-out
application is \LASKO\ over typed artifacts.

\section{A Minimal Geometric Vocabulary}

This section gives the reader a brief overview of terminology that is standard in differential geometry, which we will generalize to arbitrary tangent categories. 

A \emph{base space} $\M$ is the space of states that skill optimization searches through.  In
ordinary optimization this might be a parameter space.  In agentic skill
optimization it may be a space of typed Markdown workflows, where a point is a
particular skill artifact with prompts, schemas, validators, examples, traces,
and anchors.  A \emph{tangent vector} at a point $x\in\M$ is a first-order
direction in which that state can visibly change.  The collection of all such
directions over all points is the \emph{tangent bundle} $T\M\to\M$.  Informally,
the fiber $T_x\M$ is the local menu of visible infinitesimal changes at $x$.

A \emph{vector bundle} is the same organizational idea with a more general
fiber.  Instead of attaching visible state changes to each $x$, we attach some
linear space of local data.  For this paper the important bundle is
$\A\to\M$: the fiber $\A_x$ contains controlled edit modes available at the
skill state $x$.  These are not necessarily visible edits yet.  They may include
which optimizer policy is used, which validator route is active, which template
slot is being filled, or which rollout context is selected.  A
\emph{section} of a bundle is a rule that chooses one fiber element at every
state.  Thus a vector field is a section of $T\M$, while a skill-edit policy is
a section of $\A$.

A \emph{bundle map} is a structure-preserving map between bundles over the same
base.  The most important one here is the \emph{anchor}
\[
  \rho:\A\longrightarrow T\M.
\]
It sends a controlled edit mode to the visible first-order effect it produces
on the skill artifact.  The distinction is useful because two controlled edits
can look identical in the present document while carrying different
implementation context for later edits.  The kernel $\ker(\rho)$ is the part of
the controlled edit state that has no immediate visible effect but can still
matter downstream.

Finally, a \emph{Lie bracket} measures order-dependence between two local
directions.  If $X$ and $Y$ are visible vector fields, then $[X,Y]$ captures the
leading-order difference between ``move a little along $X$ and then $Y$'' and
``move a little along $Y$ and then $X$.''  In skill optimization, this is the
mathematical version of a common engineering fact: repairing the schema and
then rewriting the examples may not be equivalent to rewriting the examples and
then repairing the schema.  A Lie algebroid has the same idea upstairs, on
controlled edit policies:
\[
  [s,t]_\A
  \quad\text{with}\quad
  \rho([s,t]_\A)=[\rho(s),\rho(t)].
\]
The bracket says how controlled edits compose before we forget their internal
context by applying the anchor.

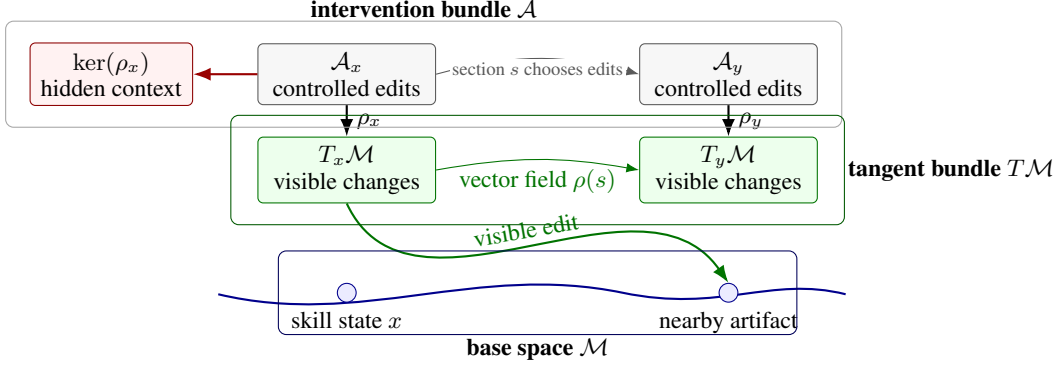
\begin{figure}[t]
  \centering
  \begin{tikzpicture}[
    font=\small,
    >=Latex,
    basept/.style={
      circle,
      draw=blue!55!black,
      fill=blue!8,
      inner sep=2.5pt
    },
    fiber/.style={
      draw=black!65,
      fill=black!3,
      rounded corners=2pt,
      minimum width=2.35cm,
      minimum height=0.68cm,
      align=center
    },
    tfiber/.style={
      draw=green!45!black,
      fill=green!7,
      rounded corners=2pt,
      minimum width=2.35cm,
      minimum height=0.68cm,
      align=center
    },
    kfiber/.style={
      draw=red!55!black,
      fill=red!5,
      rounded corners=2pt,
      minimum width=2.15cm,
      minimum height=0.62cm,
      align=center
    },
    boxlabel/.style={font=\small\bfseries, inner sep=1pt}
  ]
    \draw[blue!55!black, thick]
      (-4.2,0) .. controls (-2.3,-0.45) and (-0.8,0.42) .. (1.15,0.02)
      .. controls (2.45,-0.25) and (3.1,0.25) .. (4.1,0);
    \node[basept, label=below:skill state $x$] (x) at (-2.5,0.02) {};
    \node[basept, label=below:nearby artifact] (y) at (2.55,0.02) {};

    \node[fiber, above=2.35cm of x] (a) {$\A_x$\\controlled edits};
    \node[tfiber, above=1.05cm of x] (tx) {$T_x\M$\\visible changes};
    \node[kfiber, left=0.85cm of a] (k) {$\ker(\rho_x)$\\hidden context};

    \draw[->, thick] (a) -- node[right] {$\rho_x$} (tx);
    \draw[->, red!60!black, thick] (a) -- (k);
    \draw[->, green!45!black, thick] (tx.south) to[out=-65,in=125]
      node[above, sloped] {visible edit} (y.north);

    \node[fiber, above=2.35cm of y] (ap) {$\A_y$\\controlled edits};
    \node[tfiber, above=1.05cm of y] (ty) {$T_y\M$\\visible changes};
    \draw[->, thick] (ap) -- node[right] {$\rho_y$} (ty);

    \draw[->, black!65] (a.east) to[out=15,in=165]
      node[below, font=\scriptsize, fill=white, inner sep=1pt]
      {section $s$ chooses edits} (ap.west);
    \draw[->, green!45!black] (tx.east) to[out=10,in=170]
      node[below] {vector field $\rho(s)$} (ty.west);

    \node[draw=black!35, rounded corners=3pt, fit=(a)(ap)(k),
      inner xsep=9pt, inner ysep=8pt,
      label={[boxlabel]above:intervention bundle $\A$}]
      (abox) {};
    \node[draw=green!35!black, rounded corners=3pt, fit=(tx)(ty),
      inner xsep=10pt, inner ysep=8pt,
      label={[boxlabel]right:tangent bundle $T\M$}]
      (tbox) {};
    \node[draw=blue!35!black, rounded corners=3pt, fit=(x)(y),
      inner xsep=22pt, inner ysep=12pt,
      label={[boxlabel]below:base space $\M$}]
      (mbox) {};
  \end{tikzpicture}
  \caption{
    The geometric terms used in the paper.  The base space $\M$ contains skill
    states.  The tangent bundle $T\M$ contains visible first-order artifact
    changes.  The intervention bundle $\A$ contains controlled edit modes, and
    the anchor $\rho:\A\to T\M$ maps each controlled edit to its visible effect.
    The kernel $\ker(\rho)$ records controlled context that is not immediately
    visible but can affect later composition.
  }
  \label{fig:geometric-vocabulary}
\end{figure}

With this vocabulary, the main claim can be stated plainly.  A flat optimizer
acts as if every edit were already a visible tangent vector.  A \LASKO\ agent
uses the richer picture in \Cref{fig:geometric-vocabulary}: controlled edit
policies live in $\A$, visible artifact changes live in $T\M$, the anchor
connects the two, and brackets measure whether local repairs compose
predictably.

\section{\LASKO: Lie Algebroid SKill Optimization}

The shift from tangent bundles to Lie algebroids is not a change of metaphor;
it is the mathematical move required when an intervention has internal
structure not visible in its first-order effect.  In the flat approximation, an
intervention is simply a vector field on a model, data, or artifact manifold.
In the algebroid approximation, the intervention is a section
$s\in\Gamma(\A)$ and only its anchor $\rho(s)$ is visible as a vector field.
The difference matters whenever two edits have the same immediate artifact
effect but different future composition behavior.

\subsection{\LASKO\ and Markdown derivations}

For \LASKO, the base object is not a neural parameter vector but a typed
Markdown workflow.  The relevant algebroid is the bundle of controlled
derivations on a Markdown presheaf or workflow diagram.  A section is a possible
skill edit policy: strengthen a guardrail, add an example, repair a schema,
split a plan node, or change a validation route.  The anchor evaluates that
derivation on a concrete document or workflow and produces the visible Markdown
edit.  The kernel contains template variables, latent rubric assumptions,
unfilled slots, hidden tool preconditions, and validator-routing choices that
do not change the current artifact but alter future edit composition.

This is also the correct place to state the topological intuition carefully.
For an ordinary tangent bundle, global nonvanishing vector fields are
constrained by familiar Euler-class phenomena.  For a Lie algebroid, analogous
obstructions should be formulated through the characteristic classes or
cohomology of the algebroid and through the anchor.  Thus the agentic analogue
is not a literal hairy-ball theorem for Markdown.  It is the expectation that
nontrivial algebroid topology or cohomology can force global skill policies to
have anchor-zero states, singularities, or fixed points:
\[
  \rho(s)=0.
\]
Operationally, these are halt states, unresolved templates, validator
dead-ends, or workflow points at which a nominally global skill cannot make a
visible admissible move without adding new local structure.

\subsection{Scope of the present paper}

The same algebroid lift has natural analogues in causal discovery and neural
adapter composition, but those analogues are not developed here.  \BRIDGE\ and
\SKFM\ motivate the use of bracket residuals as diagnostics for missing
directions, while \ALLORA\ and \LICKET\ motivate bracket control in adapter
composition.  In the present paper they serve only as lineage and vocabulary.
The technical object studied below is \LASKO\ over typed workflow artifacts:
controlled sections are edit policies, the anchor is the realized artifact
change, and feedback comes from rollout traces, validators, and critiques.
\SkillOpt\ is the motivating optimizer instance, while \BASKET, \ROCKET, and
Democritus supply workflow-specific instantiations.

This boundary is important.  Algebroidic versions of \BRIDGE/\SKFM\ would ask
how to reconstruct hidden intervention directions from observed bracket
residuals.  Algebroidic versions of \ALLORA/\LICKET\ would ask how adapter
routing, rank, and constraint transport live above visible parameter updates.
Both are interesting follow-on projects, but they would require their own
observation models and experimental protocols.  The rest of this paper stays
with the \LASKO\ question: how should an agent optimize structured skills when
edits are anchored, typed, validated, and order-dependent?

\section{From Tangent-Bundle IC to Lie Algebroid IC}

\subsection{The tangent-bundle baseline}

Let $\M$ be a statistical, behavioral, or model-state manifold.  In the original
\IC\ viewpoint, an intervention $\doop(X)$ is represented locally by a vector
field $X\in \Gamma(T\M)$.  If two interventions are applied in infinitesimal
orders, their leading order order-sensitivity is measured by the Lie bracket
\[
  [X,Y] = XY - YX.
\]
For a visible family of interventions $X_1,\ldots,X_k$, write
\[
  \V = \operatorname{span}\{X_1,\ldots,X_k\}.
\]
The Frobenius diagnostic asks whether $[X_i,X_j]\in \V$ for all $i,j$.  If the
bracket leaves the visible span, then the observed intervention family is not
closed.  In a \LASKO\ setting, this non-closure means that the visible edit
library is missing some repair direction, hidden workflow condition, or
singular regime that affects later rollouts.

\subsection{Lie algebroids}

\begin{definition}[Lie algebroid]
A Lie algebroid over a smooth base $\M$ is a vector bundle $\A\to\M$ equipped
with an anchor bundle map $\rho:\A\to T\M$ and a Lie bracket
$[\,\cdot,\cdot\,]_\A$ on sections $\Gamma(\A)$ satisfying the Leibniz rule
\[
  [s_1,f s_2]_\A
  =
  f[s_1,s_2]_\A + (\rho(s_1)f)s_2
\]
for smooth functions $f\in C^\infty(\M)$ and sections $s_1,s_2\in\Gamma(\A)$.
The anchor also preserves brackets:
\[
  \rho([s_1,s_2]_\A) = [\rho(s_1),\rho(s_2)].
\]
\end{definition}

For \IC, $\A$ should be read as the bundle of \emph{available controlled
interventions}.  A section $s\in \Gamma(\A)$ is not yet a visible vector field;
it is a controlled action mode.  Its visible effect is $\rho(s)\in\Gamma(T\M)$.
This gives four objects with distinct empirical meanings:
\begin{center}
\begin{tabular}{ll}
\toprule
Object & Interpretation \\
\midrule
$s\in\Gamma(\A)$ & controlled intervention mode \\
$\rho(s)\in\Gamma(T\M)$ & visible statistical or behavioral effect \\
$[s_i,s_j]_\A$ & abstract composition bracket of interventions \\
$[\rho(s_i),\rho(s_j)]$ & visible order-sensitivity in the base \\
\bottomrule
\end{tabular}
\end{center}

\subsection{Kernel, image, and latent structure}

The kernel of the anchor is the isotropy algebra
\[
  \ker(\rho)_x = \{a\in \A_x : \rho(a)=0\}.
\]
When the rank is regular, this forms a bundle of Lie algebras.  Elements of
$\ker(\rho)$ have no immediate visible first-order effect, but they can bracket
with other sections and affect later composition.  In \LASKO\ terms, this is
latent workflow state: not merely an unobserved variable, but an unanchored
edit degree of freedom that changes the algebra of future repairs.

The image $\rho(\A)\subseteq T\M$ is the visible distribution generated by
controlled interventions.  For a genuine Lie algebroid, the full image of all
sections is closed under the ordinary Lie bracket whenever it is a smooth
distribution, because the anchor preserves brackets.  Empirical non-closure
therefore has to be interpreted carefully: it indicates non-closure of the
\emph{chosen finite visible family}, rank singularities, estimation error,
projection error, or a missing extension of the intervention algebroid.  It is
not an invariant contradiction inside an exact Lie algebroid.

This is precisely where involution algebroids are useful.  Burke and
MacAdam~\citep{burke2019involution} formulate involutivity in the setting of
tangent categories, generalizing the closure condition that ordinary Lie
algebroids impose over smooth manifolds.  For \LASKO, this supplies the
right abstraction for the question: does the finite library of visible edit
policies close under the ambient workflow tangent structure, or must the skill
be completed by adding hidden, singular, or anchor-invisible repair modes?  In
other words, the empirical closure residual estimates the gap between a
measured edit library and its involutive closure.

\begin{definition}[Visible Frobenius residual]
Let $\widehat{V}$ be the finite visible span of measured anchored fields
$\rho(s_1),\ldots,\rho(s_k)$, and let $P_{\widehat{V}}$ be a local projection.
The visible Frobenius residual is
\[
  R_{\mathrm{vis}}(i,j)
  =
  (I-P_{\widehat{V}})[\rho(s_i),\rho(s_j)].
\]
\end{definition}

\begin{definition}[Anchor defect residual]
For an estimated, projected, finite-rank, or routed intervention model
$\widehat{\A}$, define the anchor defect residual
\[
  R_{\mathrm{anchor}}(i,j)
  =
  \widehat{\rho}([s_i,s_j]_{\widehat{\A}})
  -
  [\widehat{\rho}(s_i),\widehat{\rho}(s_j)].
\]
For an exact Lie algebroid this quantity is zero by definition.  In learned
systems it is a modeling residual: it measures whether the estimated controlled
intervention algebra explains the observed visible bracket.
\end{definition}

\subsection{Regular, equivalent, and singular cases}

\LASKO decomposes the skill optimization task into three cases.  First, in the regular case $\rank(\rho)$ is locally constant.  The image of the
anchor behaves like a smooth distribution, and Frobenius-style closure can be
studied locally. Second, in equivalence or Atiyah-style cases, multiple controlled
interventions have the same anchored effect.  The kernel then represents
symmetries, redundant descriptions, conservation laws, or representation
choices.  A skill optimizer should quotient or monitor these directions before
treating every bracket residual as a substantive missing repair mode.

Third, in the singular case, $\rank(\rho)$ changes across regimes.  This is the
natural model for interventions that activate only under certain contexts:
safety policies on risky prompts, tool contracts on tool-use tasks, citation
constraints in document workflows, or evaluators active only in one task
regime.  Singular Lie algebroids therefore fit the variable-rank behavior that
is awkward in ordinary vector-space models of skill editing.

\section{The Tangent Category of Typed Markdown}

\LASKO\ operates over artifacts rather than just vectors.  We model the
base category $\Md$ as typed, anchored Markdown.  An object is a tuple
\[
  M=(A,P,\tau,\sigma)
\]
where $A$ is a finite anchor tree or poset, $P$ is a Markdown parse tree indexed
by anchors, $\tau$ assigns block types, and $\sigma$ stores semantic metadata
such as schemas, tool names, validators, provenance, or plan graph labels.
Block types include instructions, guardrails, examples, schemas, citations,
tool contracts, plan nodes, plan edges, and test cases.

A morphism $f:M\to N$ is an anchor-respecting rewrite program.  It transports
anchors, rewrites parse fragments, updates metadata, and preserves or explicitly
coerces block types.  Composition is composition of rewrite programs.

For each Markdown object $M$, let $E(M)$ be the fiber of admissible anchored
edits.  Examples include rewriting an instruction clause, strengthening a
guardrail, adding an example, changing a schema field, inserting a missing plan
action, rewiring a plan edge, adding a citation constraint, or modifying a tool
contract.  ``Infinitesimal'' means local and admissible relative to the typed
artifact, not character-level small.

\begin{definition}[Markdown tangent object]
The tangent object of $M$ is
\[
  T M = (M,E(M)).
\]
For a rewrite $f:M\to N$, the tangent map is
\[
  T f(M,\delta) = (N,Df_M(\delta)),
\]
where $Df_M$ transports anchored edits through the rewrite program.
\end{definition}

\begin{proposition}[Typed Markdown tangent structure]
Assume each edit fiber $E(M)$ is an additive commutative monoid or module over
compatible edit merge, and each rewrite $f:M\to N$ has an edit pushforward
$Df_M:E(M)\to E(N)$ preserving zero, compatible addition, identity, and
composition:
\[
  D(\operatorname{id})_M=\operatorname{id}_{E(M)},
  \qquad
  D(g\circ f)_M = Dg_{f(M)}\circ Df_M.
\]
Then typed anchored Markdown objects carry the forward structure of a tangent
category with tangent endofunctor $T$.
\end{proposition}

The proposition is intentionally conditional.  Raw Markdown strings do not have
enough structure.  The tangent category appears only after adding anchors,
types, admissible edits, and a merge or conflict policy.

Once this tangent category is fixed, the appropriate closure condition for edit
policies is not merely closure under a hand-written bracket.  It is involutive
closure internal to the tangent category.  This is the point at which
involution algebroids~\citep{burke2019involution} become the categorical
counterpart of the workflow-closure condition needed by \LASKO.

\section{Markdown Objects and Partially Ordered Workflows}

The preceding section treats a Markdown artifact as a typed object with local
anchors.  For \LASKO, and for the earlier \BASKET/\ROCKET\ plan-repair
experiments described in \emph{Categories for AGI}~\citep{mahadevan2026catagi},
the more operational object is usually not a single document but a partially
ordered workflow of document fragments, tool calls, evidence blocks, and plan
states.  This section makes that base category explicit.

\begin{definition}[Anchored Markdown object]
An anchored Markdown object is a tuple
\[
  M=(A,P,\tau,\sigma,\chi)
\]
where $A$ is a finite anchor poset, $P$ assigns Markdown syntax trees to
anchors, $\tau:A\to\mathsf{Type}$ assigns block roles, $\sigma$ stores semantic
metadata, and $\chi$ records local contracts.  Typical contracts include
schema validity, citation availability, tool preconditions, safety
constraints, evaluator hooks, and provenance obligations.
\end{definition}

The order on $A$ is not merely formatting.  It records dependency and scope.
For example, a section heading scopes its children, a tool contract scopes the
tool examples that instantiate it, a claim scopes the citations supporting it,
and a plan node scopes the tests used to validate it.  A local edit at anchor
$a$ may therefore propagate to every anchor above or below $a$ depending on the
chosen variance: changing a schema constrains children, while changing an
example may update summary metadata upward.

\begin{definition}[Workflow poset]
A workflow shape is a finite poset $(W,\leq)$ whose elements are typed stages:
document fragments, extraction steps, tool calls, validator checks, plan
states, evidence bundles, or synthesis artifacts.  The relation $u\leq v$
means that the output or contract at stage $u$ is available to, required by, or
checked before stage $v$.
\end{definition}

\begin{definition}[Markdown workflow]
A Markdown workflow is a functor
\[
  F:W\to \Md
\]
from a workflow poset to anchored Markdown objects and anchor-respecting
rewrite maps.  Thus each stage $u\in W$ has a local artifact $F(u)$, and each
dependency $u\leq v$ has a transport map
\[
  F_{uv}:F(u)\to F(v)
\]
describing how anchors, claims, contracts, or evidence flow forward through the
workflow.
\end{definition}

This definition captures the structure that \BASKET\ and \ROCKET\ were using
informally.  \BASKET\ constructs a workflow diagram from text: for example,
from a 10-K filing to extracted risks, operating mechanisms, plan fragments,
and synthesized financial narratives.  \ROCKET\ performs local search in the
neighborhood of such a diagram: repair a plan node, replace a weak evidence
edge, add a missing validator, or choose a better synthesis path.  In both
cases, the artifact is a partially ordered diagram of Markdown-like typed
objects, not a flat prompt string.

\begin{definition}[Workflow morphism]
Given workflows $F:W\to\Md$ and $G:W'\to\Md$, a morphism
$(\alpha,\eta):F\to G$ consists of an order-preserving map
$\alpha:W\to W'$ and a natural transformation
\[
  \eta_u:F(u)\to G(\alpha(u))
\]
compatible with dependency maps.  This morphism can refine a workflow, compress
several stages, rename anchors, insert validators, or translate one plan
representation into another.
\end{definition}

The category of Markdown workflows will be denoted $\mathsf{WfMd}$.  Ordinary
Markdown artifacts embed as constant or one-node workflows.  This lets the
theory handle both simple skill files and larger agentic workflows with the
same formal language.

Tangent vectors in $\mathsf{WfMd}$ are workflow-local edits.  Concretely, an
edit is a finite family
\[
  \delta = \{(u,\delta_u)\}_{u\in S\subseteq W}
\]
where each $\delta_u\in E(F(u))$ is an anchored edit and the family satisfies
dependency compatibility: if $u\leq v$, then the pushed-forward edit
$DF_{uv}(\delta_u)$ must either agree with $\delta_v$, be absorbed by the
contract at $v$, or be recorded as a boundary residual.  This is the operational
origin of curvature in plan repair.  Two edits may commute on isolated Markdown
blocks but fail to commute once their effects are transported through the
workflow poset.

Partially ordered workflows also explain why colimits and limits keep
reappearing in this program.  Forward execution aggregates local artifacts into
a user-facing synthesis, a colimit-like operation over the workflow diagram.
Backward critique restricts a global failure to the stages and anchors that
could have caused it, a limit-like or cotangent operation.  \DB\ is the
backward transport of this critique signal through $W$, while \ROCKET-style
repair chooses the next tangent direction in $\mathsf{WfMd}$.

\section{Reverse Skill Signals and Diagrammatic Backpropagation}

Forward tangent structure explains how a proposed edit moves through a skill
workflow.  \LASKO\ also needs a backward pass.  After rollout, the system
observes validator failures, critic comments, safety rejections, bad tool
calls, or weak held-out scores.  These signals are not useful until they are
localized: the optimizer must decide whether the repair belongs at a schema
field, a guardrail, an example, a plan node, a tool contract, or a validator.

We model this localization by a reverse tangent or cotangent structure over the
workflow category:
\[
  T^*F \to F.
\]
An element of $T^*F$ is an anchored reverse skill signal: a blame assignment,
critique gradient, validation residual, score delta, or loss annotation attached
to stages and anchors of the workflow.  For a workflow morphism
$f:F\to G$, the reverse map pulls a downstream signal back to possible upstream
causes,
\[
  T^*f : T^*G \to T^*F,
\]
and satisfies the reverse chain rule
\[
  T^*(g\circ f) = T^*f \circ T^*g.
\]
This is the \LASKO\ analogue of backpropagation.  The objects are not tensors
alone; they are skill cards, plans, tool contracts, examples, traces, and
validators connected by workflow dependencies.

\begin{definition}[Diagrammatic backpropagation curvature]
Let $X$ and $Y$ be two anchored edit fields on a workflow $F$.  Let
$\eta_{XY}$ be the reverse skill signal obtained by applying $X$ then $Y$,
running the resulting workflow, and pulling the validation signal back to
anchors.  Define $\eta_{YX}$ analogously for the opposite edit order.  A
diagrammatic backpropagation curvature diagnostic is
\[
  \Curv_{\D}(X,Y)
  =
  d(\eta_{XY},\eta_{YX}),
\]
where $d$ is a chosen distance between anchored reverse signals.
\end{definition}

This curvature measures path-dependence of credit assignment.  Two edit paths
can produce similar final Markdown while inducing different explanations of why
a validator passed or failed.  Such mismatches are exactly what make
long-horizon skill repair unstable: the optimizer receives a local signal, but
the signal depends on the order in which previous repairs exposed, masked, or
rerouted the relevant anchors.

\section{\LASKO\ as Lie Algebroid Optimization}

The tangent object $T\mathsf{WfMd}$ contains visible anchored workflow edits.
A real \LASKO\ instance has a smaller and more structured family of controlled edit policies:
repair a schema, strengthen a guardrail, add a counterexample, localize a tool
contract, split a plan node, rewrite a rubric, or insert a citation constraint.
Thus the natural object is a Lie algebroid
\[
  \A \to \mathsf{WfMd},
  \qquad
  \rho:\A\to T\mathsf{WfMd},
  \qquad
  [\,\cdot,\cdot\,]_\A.
\]
The fiber $\A_M$ is the space of controlled intervention modes available at a
workflow object $M$.  The anchor $\rho(a)$ is the visible anchored workflow
edit.  The kernel $\ker(\rho)$ contains hidden procedural structure: unstated
rubrics, implicit tool assumptions, latent safety policies, template variables,
or unfilled slots that do not affect the current execution but change
composition.

\begin{definition}[\LASKO\ residuals]
Given controlled edit sections $s_i,s_j\in\Gamma(\A)$, define
\[
\begin{aligned}
  R_{\mathrm{Md}}(i,j)
    &= (I-P_{\widehat{V}})[\rho(s_i),\rho(s_j)],\\
  R_{\mathrm{ker}}(i,j)
    &= \pi_{\ker(\rho)}([s_i,s_j]_\A),\\
  R_{\mathrm{DB}}(i,j)
    &= \Curv_{\D}(\rho(s_i),\rho(s_j)).
\end{aligned}
\]
These measure visible edit nonclosure, hidden procedural interaction, and
path-dependent reverse blame transport.
\end{definition}

The combined causal curvature loss is
\[
\begin{aligned}
  \Loss_{\text{\LASKO}}
  =
  \Loss_{\mathrm{task}}
  &+ \lambda_{\mathrm{IC}}\sum_{i<j}\|R_{\mathrm{Md}}(i,j)\|^2 \\
  &+ \lambda_{\mathrm{ker}}\sum_{i<j}\|R_{\mathrm{ker}}(i,j)\|^2 \\
  &+ \lambda_{\mathrm{DB}}\sum_{i<j}\|R_{\mathrm{DB}}(i,j)\|^2 \\
  &+ \lambda_{\mathrm{A}}\sum_{i,j,k}\|R_\A(s_i,s_j)s_k\|^2,
\end{aligned}
\]
where $R_\A$ is the curvature of a chosen connection on the intervention
algebroid.  The connection specifies how controlled edit modes transport across
changed documents:
\[
  R_\A(s_i,s_j)s_k
  =
  \nabla_i\nabla_j s_k
  -
  \nabla_j\nabla_i s_k
  -
  \nabla_{[s_i,s_j]_\A}s_k.
\]

\begin{conjecture}[Low-curvature skill fixed points]
Let $\Phi$ be the combined forward rollout, reverse critique, and edit-selection
operator of a \LASKO\ system on a typed Markdown tangent category.  Under
compactness, continuity, and bounded-edit assumptions appropriate to a chosen
edit theory, locally optimal stable skills occur at fixed points of $\Phi$ where
the task gradient, visible IC residual, kernel interaction residual, DB
curvature, and algebroid connection curvature vanish relative to the observation
functor.
\end{conjecture}

This is a categorical analogue of a Bellman optimality condition.  Value
iteration becomes repeated forward-backward transport in a tangent category.
Convergence is obstructed by causal curvature: noncommuting edits, hidden
procedural interactions, and path-dependent reverse blame.

\section{Experimental Program: \LASKO\ over Workflow Algebroids}

The most direct way to test the framework is not to begin with an abstract
algebroid reconstruction algorithm, but with a small \LASKO\ experiment in
which all objects are inspectable.  The base object is a Markdown workflow
$F:W\to\Md$ built from a short task document, a tool contract, a few examples,
and one or more validators.  The intervention algebroid is generated by a
finite set of named edit policies:
\[
  s_{\mathrm{schema}},\quad
  s_{\mathrm{guardrail}},\quad
  s_{\mathrm{example}},\quad
  s_{\mathrm{tool}},\quad
  s_{\mathrm{plan}}.
\]
Each section is a controlled edit mode.  The anchor sends it to the visible
workflow edit that is actually applied:
\[
  \rho(s_i)\in T_F\mathsf{WfMd}.
\]
The experiment then compares ordered edit pairs.  Starting from the same
workflow $F$, apply $s_i$ followed by $s_j$, and compare this with $s_j$
followed by $s_i$.  The final artifacts, validator results, and reverse blame
signals give empirical estimates of three quantities:
\[
\begin{aligned}
  R_{\mathrm{vis}}(i,j)
    &= d_{\mathrm{art}}(F_{ij},F_{ji}),\\
  R_{\mathrm{DB}}(i,j)
    &= d_{\mathrm{blame}}(\eta_{ij},\eta_{ji}),\\
  R_{\mathrm{ker}}(i,j)
    &= d_{\mathrm{trace}}(\theta_{ij},\theta_{ji})
       \quad\text{when } d_{\mathrm{art}}(F_{ij},F_{ji})\approx 0.
\end{aligned}
\]
Here $d_{\mathrm{art}}$ measures visible artifact difference, $d_{\mathrm{blame}}$
measures the path-dependence of validator blame, and $d_{\mathrm{trace}}$
measures hidden procedural differences such as changed tool assumptions,
different anchor provenance, or different validator routes.  The last term is
a practical proxy for kernel energy: the visible artifact may look unchanged,
but the workflow trace shows that the controlled edit algebra has changed.

\subsection{A minimal \LASKO\ benchmark}

A first benchmark can use three small workflow families.

\begin{center}
\begin{tabular}{p{0.25\linewidth}p{0.62\linewidth}}
\toprule
Workflow family & Example repair task \\
\midrule
Skill card & Repair a tool-using Markdown skill with schema, guardrails, examples, and tests. \\
Plan graph & Repair a partially ordered plan with missing preconditions and inconsistent evidence edges. \\
10-K workflow & Repair a \BASKET/\ROCKET-style extraction workflow from filing fragments to risks, mechanisms, and narratives. \\
\bottomrule
\end{tabular}
\end{center}

Each workflow is initialized with seeded defects: a missing schema field, a
misplaced guardrail, an example that contradicts the tool contract, a plan edge
that bypasses evidence, or a validator that fires at the wrong stage.  The
system runs a small library of edit policies in all ordered pairs.  For each
pair it records final validator score, visible artifact distance, reverse blame
locality, and trace-level disagreement.  The central hypothesis is:
\[
  \text{higher curvature residuals predict worse repair stability.}
\]
In operational terms, edit pairs with large $R_{\mathrm{vis}}$ or
$R_{\mathrm{DB}}$ should be more likely to produce validator regressions,
diffuse blame assignment, or unstable follow-up edits.  Pairs with small
$R_{\mathrm{vis}}$ but large $R_{\mathrm{ker}}$ are especially interesting:
they are candidates for latent workflow structure, because their visible
Markdown outputs agree while their traces or future compositions disagree.

\subsection{What counts as algebroid structure?}

The experiment does not need a perfect symbolic Lie algebroid.  It needs an
estimated algebroid with four observable parts:
\begin{enumerate}[leftmargin=*]
  \item a finite library of sections, the named edit policies;
  \item an anchor, the realized workflow edit emitted by each policy;
  \item a bracket estimator, computed by order-contrast of edit pairs;
  \item a kernel proxy, obtained from trace disagreement when visible artifact
        distance is small.
\end{enumerate}
This is enough to test whether the algebroid vocabulary earns its keep.  If the
kernel proxy predicts future order-sensitivity, then the anchor/kernel
distinction is not merely decorative: it identifies hidden procedural state
that ordinary Markdown diffing misses.

\subsection{Ten-anchor controlled benchmark}

The cleanest empirical setting is a controlled benchmark in which the
interaction structure is known.  We therefore construct a ten-anchor Markdown
skill chain with anchors
\[
\begin{gathered}
\texttt{schema},\ \texttt{normalization},\ \texttt{evidence},\
\texttt{citation},\ \texttt{abstention},\ \texttt{format},\\
\texttt{tool\_contract},\ \texttt{routing},\ \texttt{validator},\
\texttt{final\_answer}.
\end{gathered}
\]
There are ten corresponding edit sections.  Five ordered pairs are productive:
\[
\begin{array}{lll}
\texttt{schema}\to\texttt{normalization}, &&
\texttt{evidence}\to\texttt{citation},\\
\texttt{abstention}\to\texttt{format}, &&
\texttt{tool\_contract}\to\texttt{routing},\\
\texttt{validator}\to\texttt{final\_answer}. &&
\end{array}
\]
Reversing a productive pair loses the benefit.  In other words, the benchmark
is deliberately designed so that useful repair directions live in ordered edit
interactions rather than in independent single edits.  This is the point at
which ordinary \SkillOpt\ begins to resemble a combinatorial search over edit
sequences.

We compare four strategies.  \emph{Greedy single-edit} search evaluates local
one-edit candidates, as in a short-horizon \SkillOpt\ loop with edit budget
one.  \emph{Exhaustive ordered pairs} evaluates all $10\cdot 9=90$ ordered edit
pairs.  \emph{Random ordered pairs} samples ten ordered pairs.  The
\emph{LASKO-prioritized} method first computes a cheap bracket proxy from
the anchor/order structure and then validates only the ten highest-bracket
ordered pairs.  Bracket probes are algebraic diagnostics; validation probes are
the expensive rollout/gate calls.

\begin{center}
\begin{tabular}{lccc}
\toprule
Method & Bracket probes & Validation probes & Score \\
\midrule
Greedy single-edit & 0 & 27 & 0.240 \\
Random ordered pairs & 0 & 10 & 0.266 mean / 0.800 max \\
Exhaustive ordered pairs & 0 & 90 & 1.000 \\
LASKO-prioritized & 90 & 10 & 1.000 \\
\bottomrule
\end{tabular}
\end{center}

We then wrapped the same controlled task in the local \SkillOpt\ serving
contract.  Each candidate edit sequence is rendered as a Markdown skill and
sent to an Exo-served OpenAI-compatible chat endpoint running
\texttt{mlx-community/Llama-3.2-3B-Instruct-4bit}.  The score is progress
times faithfulness: progress counts closed productive chains, while
faithfulness checks whether the served model reads the skill state correctly.
The fully Exo-served run gives:
\begin{center}
\begin{tabular}{lccc}
\toprule
Method & Bracket probes & Exo validation probes & Score \\
\midrule
Greedy single-edit & 0 & 10 & 0.000 \\
Random ordered pairs & 0 & 11 & 0.120 \\
Exhaustive ordered pairs & 0 & 91 & 1.000 \\
LASKO-prioritized & 90 & 11 & 1.000 \\
\bottomrule
\end{tabular}
\end{center}

We also scale the construction by increasing the number of independent
productive chains.  With $k$ productive chains there are $N=2k$ edits and
$N(N-1)$ ordered edit pairs.  Exhaustive validation therefore grows
quadratically in $N$, while the bracket screen validates only the top $2k$
ordered pairs plus a small number of focused final validation chunks.  The
same Exo-served target model remains faithful through the $160$-edit setting
when validation is focused by chain subset:
\begin{center}
\begin{tabular}{lccc}
\toprule
Edits & Exhaustive Exo probes & LASKO Exo probes & Score \\
\midrule
10 & 91 & 11 & 1.000 \\
20 & 381 & 21 & 1.000 \\
40 & 1561 & 42 & 1.000 \\
80 & 6321 & 84 & 1.000 \\
160 & 25441 & 168 & 1.000 \\
\bottomrule
\end{tabular}
\end{center}

This is the intended analogue of the ten-node latent-confounded chain used in
the \BRIDGE/\SKFM\ setting.  The win is not that the controlled benchmark is
realistic; it is that the failure mode is isolated.  Short-horizon greedy
\SkillOpt\ sees only small local gains and recovers only part of the interaction
chain.  Exhaustive ordered-pair search recovers the full chain but pays the
enumeration cost.  The LASKO-prioritized method recovers the exhaustive
score while spending a linear number of focused validation probes after the
bracket screen.  The high-level message is therefore specific: when skill edits
do not commute, the algebroid bracket is a screening device for edit
interactions, reducing expensive validation from brute-force sequence
enumeration to targeted pair probing.

\subsection{Agent-workflow controlled benchmark}

The ten-anchor chain establishes the validation-economics claim, but its
anchors are intentionally schematic.  We therefore added a second controlled
benchmark whose anchors are recognizable agent workflow contracts:
\[
\begin{gathered}
\texttt{task\_contract},\ \texttt{plan\_graph},\
\texttt{tool\_manifest},\ \texttt{router\_policy},\
\texttt{observation\_schema},\\
\texttt{normalization\_rules},\ \texttt{evidence\_ledger},\
\texttt{citation\_policy},\ \texttt{answer\_template},\
\texttt{validator\_contract},\\
\texttt{failure\_triage},\ \texttt{handoff\_memory}.
\end{gathered}
\]
Six workflow scenarios reuse these contracts: retrieval QA, data extraction,
code patching, research synthesis, long-horizon agency, and tool incident
triage.  A scenario succeeds only when its required repairs occur in the right
order.  For example, citation binding depends on the evidence ledger,
tool-routing repair depends on tool-affordance repair, and failure triage
depends on the validator contract.  This makes the benchmark closer to a
\LASKO\ workflow loop: edits are not merely local prompt clauses but repairs
to staged agent infrastructure.

The bracket proxy is computed from cheap workflow metadata.  Each edit section
has read and write anchors; a pair receives high bracket score when the first
edit writes an anchor read by the second, when the reverse order is also
plausible, or when the pair appears as a scenario-local ordered dependency.
Expensive validation then scores only the selected ordered pairs and the final
linearized workflow.
\begin{center}
\begin{tabular}{lccc}
\toprule
Method & Bracket probes & Validation probes & Mean workflow score \\
\midrule
Greedy single-edit & 0 & 12 & 0.000 \\
Random ordered pairs & 0 & 17 & 0.322 mean / 0.860 max \\
Exhaustive ordered pairs & 0 & 133 & 1.000 \\
LASKO-prioritized & 132 & 17 & 1.000 \\
\bottomrule
\end{tabular}
\end{center}
The LASKO-prioritized method recovers the exhaustive workflow repair while
using the same expensive validation budget as the random baseline.  The
difference is that the bracket screen uses anchor/read-write structure to buy
the validations that correspond to real workflow dependencies.  This is a
better operational picture of agentic skill optimization than the anonymous
anchor chain: the objects being repaired are the things an agent actually uses
to act, observe, validate, recover, and hand off state.

\subsection{10-K BASKET/ROCKET workflow benchmark}

The next controlled instantiation uses a Democritus-style financial research
workflow.  The anchors are now BASKET/ROCKET artifacts rather than generic
agent contracts:
\[
\begin{gathered}
\texttt{section\_router},\ \texttt{business\_model\_claims},\
\texttt{risk\_factor\_rebuttals},\ \texttt{mda\_operating\_claims},\\
\texttt{numeric\_table\_extractor},\ \texttt{numeric\_table\_gate},\
\texttt{evidence\_ledger},\ \texttt{citation\_binding},\\
\texttt{mechanism\_graph},\ \texttt{macro\_motif\_completion},\
\texttt{financial\_outcome\_alignment},\\
\texttt{temporal\_stability},\ \texttt{narrative\_synthesis},\
\texttt{report\_validator}.
\end{gathered}
\]
Seven scenarios model common 10-K research paths: business model to revenue,
risk-adjusted thesis construction, numeric financial claims, MD\&A mechanism
repair, cross-year ROCKET stability, Democritus report gating, and the full
BASKET/ROCKET loop.  The ordered dependencies are domain-specific.  A numeric
table gate must follow table extraction before a verified claim enters the
evidence ledger; risk rebuttals must be extracted before a risk-adjusted
narrative; ROCKET mechanism graphs must precede macro motif completion and
financial outcome alignment; temporal stability must be checked before final
narrative synthesis.

\begin{center}
\begin{tabular}{lccc}
\toprule
Method & Bracket probes & Validation probes & Mean 10-K score \\
\midrule
Greedy single-edit & 0 & 14 & 0.000 \\
Random ordered pairs & 0 & 29 & 0.525 mean / 1.000 max \\
Exhaustive ordered pairs & 0 & 183 & 1.000 \\
LASKO-prioritized & 182 & 29 & 1.000 \\
\bottomrule
\end{tabular}
\end{center}

We also ran this 10-K workflow through an Exo-hosted
\texttt{mlx-community/gpt-oss-20b-MXFP4-Q8} model.  Each candidate sequence was
rendered as a compact workflow skill; the served model returned the
scenario-scoped edge IDs it judged closed, and the script converted those
closed edges into scenario scores.  In this live served-model run, exhaustive
ordered-pair validation again reached score $1.000$ with $183$ Exo validation
probes, while LASKO-prioritized validation reached the same score with
$29$ Exo validation probes after $182$ cheap bracket probes.  A same-budget
random ordered-pair run scored $0.537$ under the served model and $0.439$ under
the oracle, with served-model faithfulness $0.903$.

This benchmark is still controlled, but it is no longer anonymous.  It says how
the bracket screen would be used inside a real BASKET/ROCKET repair loop:
validate high-bracket pairs such as table extraction before numeric gating,
business claims before mechanism graphs, evidence ledgers before citations, and
macro motif completion before outcome alignment.  It also gives a clean bridge
to the existing ROCKET trace artifacts: future runs can replace the synthetic
scenario graph with observed \texttt{base\_plan}\(\to\)\texttt{selected\_plan}
deltas and use the same algebroid validation accounting.

\subsection{Trace-derived Democritus LASKO benchmark}

The 10-K workflow benchmark still uses a hand-specified scenario graph.  As a
first step toward real LASKO validation, we next use completed
Prometheus/Democritus runs as the source of the workflow graph itself.  A
Democritus run exposes an ordered agent trace and its materialized artifacts:
query planning, corpus materialization, document intake, topic graph
construction, causal question generation, causal statement extraction,
relational triple extraction, CSQL materialization, PSR/manifold construction,
Prometheus claim-world gluing, and report/dashboard validation.  These stages
become the repair anchors, and observed artifact dependencies become the
required ordered edges.

We instantiated this benchmark on two Prometheus runs.  The first is a focused
physical-exercise run for the claim that regular physical exercise improves
cardiovascular health and reduces heart-disease risk.  It contains three
document episodes, $380$ CSQL claims, $47$ CSQL domains, and three PSR
episodes.  The second is a larger GLP-1 run over recent weight-loss-drug
studies.  It contains eleven document episodes, $3376$ CSQL claims, $190$ CSQL
domains, and eleven PSR episodes.  In both cases the benchmark verifies the
presence of the expected artifact surfaces: query plans, selected documents,
document manifests, topic graphs, causal questions, causal statements,
relational triples, CSQL stores, PSR Hankel bundles, Prometheus world models,
persistent states, reports, and claim atlases.

\begin{center}
\begin{tabular}{lcccc}
\toprule
Case & Method & Bracket probes & Validation probes & Mean LASKO score \\
\midrule
Physical exercise & Greedy single-edit & 0 & 11 & 0.000 \\
Physical exercise & Exhaustive ordered pairs & 0 & 111 & 1.000 \\
Physical exercise & LASKO-prioritized & 110 & 21 & 1.000 \\
GLP-1 studies & Greedy single-edit & 0 & 11 & 0.000 \\
GLP-1 studies & Exhaustive ordered pairs & 0 & 111 & 1.000 \\
GLP-1 studies & LASKO-prioritized & 110 & 21 & 1.000 \\
\bottomrule
\end{tabular}
\end{center}

We then repeated both cases with served validation through a local
\texttt{mlx-community/gpt-oss-20b-MXFP4-Q8} endpoint.  The served validator
receives the candidate repair sequence, the Democritus artifact summary, and
the allowed LASKO edge catalog, and must return strict JSON containing only the
closed ordered edges.  The physical-exercise and GLP-1 runs each produced $54$
served evaluation records with zero JSON parse failures.  In both served runs,
greedy single-edit validation remains at score $0.000$, random ordered-pair
validation reaches score $0.700$, and the LASKO-prioritized policy reaches
score $1.000$ with $21$ served validation probes and faithfulness $1.000$ to
the oracle edge computation.  Exhaustive ordered-pair validation is retained as
the deterministic oracle reference at $111$ probes.  The validation topology is
the same in the two cases, but the artifact substrates are different by nearly
an order of magnitude in extracted claims.  This matters for the LASKO story:
the algebroid screen is not tied to a toy document.  It can be applied to real
Prometheus/Democritus run folders, where the workflow stages are explicit, the
artifacts are inspectable, and expensive validation can be focused on the
high-bracket repair pairs.

\subsection{PSR-LASKO: repairing a causal manifold}

The previous Democritus cases treat the run as an ordered agent workflow.  A
more intrinsic PSR-LASKO view treats the Democritus output itself as the base
object.  The object is a predictive-state causal manifold
\[
  \mathcal M_{\mathrm{psr}}=(E,C,H,T,P,W,R,G),
\]
where \(E\) is the event stream, \(C\) the family of context charts, \(H\) and
\(T\) the local histories and tests, \(P_c(h,t)\) the local predictive cells,
\(W\) the claim--test witnesses, \(R\) the restriction maps, and \(G\) the
glued causal manifold.  Sections of the corresponding algebroid
\[
  A_{\mathrm{psr}}\longrightarrow \mathcal M_{\mathrm{psr}}
\]
are local repairs such as scoping the query-relevant manifold, splitting or
merging domain charts, canonicalizing causal entities, adding PSR tests,
pruning histories, relinking witnesses, introducing mediators, repairing
restriction gluing, and routing diagrammatic-backpropagation blame.  The
anchor sends each abstract repair to its visible effect on events, Hankel
cells, witnesses, restriction diagnostics, and the final glued manifold.

We instantiated this PSR-LASKO benchmark on a Washington Post coffee-benefits
Democritus run.  The run contains one PSR episode, nine contexts, \(346\)
events/relational triples, \(200\) causal statements, \(18\) PSR test
witnesses, and \(8/8\) compatible restriction checks.  It is a useful stress
case because only \(33.5\%\) of the extracted events lie in coffee-relevant
domains, while \(65.6\%\) lie in drifted but structurally similar
neighborhoods such as advertising, Parkinson's disease, alcohol, finance, and
generic lower-risk reasoning.  Thus the problem is not merely extracting local
triples about coffee; it is constructing the right causal manifold and
quarantining off-query charts.

\begin{center}
\begin{tabular}{lcccc}
\toprule
Method & Bracket probes & Validation probes & Score & Mean residual \\
\midrule
Greedy repair & 0 & 45 & 0.339 & 0.702 \\
Brute-force ordered pairs & 0 & 73 & 1.000 & 0.000 \\
PSR-LASKO bracket screening & 72 & 19 & 1.000 & 0.000 \\
Random same-budget pairs & 0 & 19 & 0.645 & -- \\
\bottomrule
\end{tabular}
\end{center}

The residual vector has six operational components: Hankel residual
\(R_H\), restriction residual \(R_{\mathrm{res}}\), witness residual \(R_W\),
causal-sign/mediator residual \(R_{\mathrm{causal}}\), query-scope residual
\(R_{\mathrm{scope}}\), and DB blame residual \(R_{\mathrm{DB}}\).  Brute force
and PSR-LASKO find the same zero-residual repair program:
\[
\begin{array}{l}
s_{\mathrm{scope}}\to s_{\mathrm{context}}\to s_{\mathrm{canon}}\to
s_{\mathrm{test}}\to s_{\mathrm{history}}\to s_{\mathrm{witness}}\\
\qquad\to s_{\mathrm{mediator}}\to s_{\mathrm{glue}}\to s_{\mathrm{DB}}.
\end{array}
\]
The difference is validation economics.  Brute force validates every ordered
pair of PSR sections, whereas PSR-LASKO uses \(72\) cheap bracket probes to
focus validation on \(19\) candidate pairs and the final manifold.  This is the
causal-PSR analogue of the crossword example: local causal triples are clue
answers, context charts are grid regions, shared PSR tests are crossing
letters, and the target object is not a bag of plausible triples but a single
globally glued causal manifold.

We then repeated the PSR-LASKO construction on a product brand-feedback
foundry generated by Prometheus from the query ``How comfortable is the
Lovesac sectional sofa?''  This is a different substrate from the Democritus
coffee case: the input is a review-oriented product feedback run, so the
sections are product-review repairs rather than scientific causal repairs:
scope the comfort question, normalize review sources, extract
aspect-sentiment evidence, build usage contexts, construct product tests,
stabilize purchase histories, relink review witnesses, repair brand gluing,
and route return-risk blame.  The materialized PSR object contains \(5\)
review episodes, \(53\) product-review events, \(7\) contexts, \(124\) PSR
test witnesses, and \(5/6\) compatible restriction checks.  The event stream
is deliberately mixed: \(32.1\%\) of events are product-relevant and
\(21.1\%\) carry return-risk or negative-signal pressure.

\begin{center}
\begin{tabular}{lcccc}
\toprule
Method & Bracket probes & Validation probes & Score & Mean residual \\
\midrule
Greedy repair & 0 & 45 & 0.615 & -- \\
Brute-force ordered pairs & 0 & 73 & 1.000 & 0.000 \\
PSR-LASKO bracket screening & 72 & 19 & 1.000 & 0.000 \\
Random same-budget pairs & 0 & 19 & 0.628 & -- \\
\bottomrule
\end{tabular}
\end{center}

The local Exo-served run used the same OpenAI-compatible validation contract
as the coffee experiment, with a larger token budget because the local model
otherwise clipped verbose edge validation.  Brute force again obtains score
\(1.000\) after \(73\) served validation probes.  PSR-LASKO obtains the same
score using \(72\) cheap bracket probes and \(19\) served validation probes.
The served validations took \(212.396\) seconds in total, or \(11.179\)
seconds per call on average; the bracket screen took \(0.000127\) seconds in
total.  The product result is therefore not just another Democritus trace: it
shows that the same algebroid routing primitive transfers to a PSR foundry
whose anchors are reviews, aspect sentiment, usage contexts, witnesses, brand
gluing, and purchase-risk diagnostics.

\subsection{SearchQA gated sanity check}

As a sanity check, we also ran the native \SkillOpt\ SearchQA environment rather
than a bespoke workflow.  Both optimizer and target are a locally hosted
\texttt{mlx-community/Llama-3.2-3B-Instruct-4bit} model served through an
OpenAI-compatible Exo endpoint.  The run uses the standard
rollout--reflect--merge--rank--validate loop with one epoch, eight training
items, eight held-out selection items, batch size four, edit budget one, and the
\SkillOpt\ validation gate enabled.  This is intentionally a small diagnostic,
not the main benchmark claim: its purpose is to check whether native \SkillOpt\ traces
already expose algebroid-relevant structure.

\begin{center}
\begin{tabular}{lcccccc}
\toprule
Mode & Seed & Baseline & Best & Accept & Reject & Tokens \\
\midrule
Force accept & 424344 & 0.375 & 0.500 & 2 & 0 & 56,273 \\
Gated & 424344 & 0.375 & 0.500 & 1 & 1 & 56,247 \\
\bottomrule
\end{tabular}
\end{center}

The gated run completed in 19 seconds and accepted the first candidate while
rejecting the second.  The accepted edit was a success-derived Markdown section,
\texttt{Royal Succession Rules}, with support count four; it improved selection
hard accuracy from $0.375$ to $0.500$.  The rejected second edit was also
success-derived and topical, describing a miniseries theme, and fell back to the
baseline hard score $0.375$.  Thus vanilla \SkillOpt\ supplies a necessary
native baseline: the gate prevents one local degradation, while the trace
records a rejected visible move.  By itself, however, this does not yet
establish the algebroid case.  It only shows that the native trace has enough
structure to define the next test.

More importantly for the present paper, the accepted improvement is not fully
explained by the semantic content of the accepted section.  The best-step item
flip was a Christmas-carol question whose answer changed from
\texttt{Debbie} to \texttt{FUM}; the learned section about royal succession does
not directly encode this repair.  This is precisely the kind of phenomenon the
algebroid view is meant to isolate.  The visible anchor records a successful
held-out move, but the controlled section is a brittle topical edit whose
future composability is uncertain.  The next empirical step is therefore to
mine these traces for section labels, anchor validity, rejected local moves, and
order-sensitive edit pairs.

The algebroid-facing probe is a counterfactual \emph{edit fiber} over a fixed
pre-skill state.  At each \SkillOpt\ step, keep the same current skill and
materialize every merged edit as a separate candidate section before ranking.
For the gated SearchQA trace above, step 2 contains two such neighboring
sections: a failure-derived date/answer-formatting repair with support count
three, and a success-derived miniseries-theme edit with support count one.  The
ranker selected the success-derived edit, and the validation gate rejected the
result.  Evaluating these sibling sections independently on the same held-out
selection set is the first direct algebroid diagnostic: it compares nearby
controlled sections with the same base point, the same anchor evaluation
protocol, and different semantic provenance.

\begin{center}
\begin{tabular}{lccccc}
\toprule
Step/edit & Selected & Source & Support & Hard & Soft \\
\midrule
1.1 & yes & success & 4 & 0.500 & 0.522 \\
2.1 & no & failure & 3 & 0.375 & 0.443 \\
2.2 & yes & success & 1 & 0.375 & 0.395 \\
\bottomrule
\end{tabular}
\end{center}

The result is not yet decisive: the unselected failure-derived edit does not
beat the selected topical edit on hard accuracy.  But it does dominate it on
the soft score under the same anchor evaluation.  This is a small but concrete
fiber-level signal.  The ranker selected a success-derived topical section,
while a neighboring failure-derived section at the same base point carried
higher support and better soft transfer.  The algebroid claim becomes stronger
if this pattern persists across seeds or if ordered pairs of such sections show
path-dependent validation behavior.

\subsection{ALFWorld micro-benchmark: closing one live repair edge}

We also ran a tiny native \SkillOpt/ALFWorld probe to connect the controlled
edge-ID experiments to a live embodied environment.  The purpose is deliberately
narrow.  It is not a broad ALFWorld result; it is a LASKO-style micro-benchmark
showing that a failed rollout can be localized to an ordered repair edge, that
a local skill repair can target that edge, and that the next live rollout can
close it.

The task is a single \texttt{look\_at\_obj\_in\_light} training episode:
\[
  \texttt{look at alarmclock under the desklamp.}
\]
The local Exo endpoint served \texttt{mlx-community/gpt-oss-20b-MXFP4-Q8}.
An oracle check in the same simulator solves the instance with four actions:
\[
\begin{array}{l}
\texttt{go to desk 1}\\
\texttt{take alarmclock 3 from desk 1}\\
\texttt{go to dresser 1}\\
\texttt{use desklamp 1}.
\end{array}
\]
The last action gives reward $10.0$, terminates the episode, and sets
\texttt{won=true}.  Thus the key repair is not generic exploration or final
examination.  After picking up the target object, the policy must navigate to
the light source before using it.

We encode the local edge catalog as
\[
\begin{aligned}
\texttt{look\_at\_obj\_in\_light.e1}:&
\quad \texttt{repair\_goal\_parse}
      \to \texttt{repair\_visible\_object\_pickup},\\
\texttt{look\_at\_obj\_in\_light.e2}:&
\quad \texttt{repair\_visible\_object\_pickup}
      \to \texttt{repair\_toggle\_target\_navigation},\\
\texttt{look\_at\_obj\_in\_light.e3}:&
\quad \texttt{repair\_toggle\_target\_navigation}
      \to \texttt{repair\_toggle\_target\_use},\\
\texttt{look\_at\_obj\_in\_light.e4}:&
\quad \texttt{repair\_toggle\_target\_use}
      \to \texttt{repair\_completion\_check}.
\end{aligned}
\]
Three live rollouts then test increasingly specific repairs.

\begin{center}
\begin{tabular}{lcccc}
\toprule
Repair tested & Hard & Turns & Fallbacks & First post-pickup navigation \\
\midrule
\texttt{repair\_light\_examination} & 0 & 12 & 5 & -- \\
\texttt{repair\_toggle\_target\_navigation} & 0 & 20 & 10 & \texttt{go to shelf 1} \\
\texttt{repair\_lamp\_location\_search\_order} & 1 & 5 & 1 & \texttt{go to dresser 1} \\
\bottomrule
\end{tabular}
\end{center}

The first repair finds and manipulates the visible alarm clock but does not
reach the lamp.  The second repair correctly says to search for a toggle
target, but the live policy chooses \texttt{go to shelf 1} and falls into a
desk/shelf loop.  The third repair adds a single ordering rule: after taking
the target object, prefer \texttt{go to dresser ...} before
\texttt{go to shelf ...} when searching for a desklamp.  The resulting live
trajectory is
\[
\begin{array}{l}
\texttt{go to desk 1}\\
\texttt{take alarmclock 2 from desk 1}\\
\texttt{look}\\
\texttt{go to dresser 1}\\
\texttt{use desklamp 1},
\end{array}
\]
which succeeds with hard and soft score $1.0$.

This is the smallest useful embodied version of the paper's claim.  The
controlled algebroid story says that useful directions are often not isolated
edits but ordered repair edges.  The ALFWorld probe observes exactly such an
edge in a native environment:
\[
  \texttt{repair\_visible\_object\_pickup}
  \to
  \texttt{repair\_toggle\_target\_navigation}.
\]
A generic lamp rule does not close it; a targeted navigation-order repair does.
The result therefore serves as a live sanity check for the LASKO search
primitive before scaling to broader ALFWorld suites.

\subsection{Preliminary diagnostic: Odyssey causal-claim \SkillOpt}

An initial offline diagnostic was run on an \texttt{odyssey\_causal\_claim}
\SkillOpt\ trace.  The task is to repair a Markdown skill for admitting,
rejecting, or qualifying causal claims extracted from Democritus/Odyssey case
studies, using the local truth-preserving foundry framework of
\citet{mahadevan2026odyssey}.  Candidate edits are grouped into section components such as
\texttt{claim\_extraction}, \texttt{evidence\_grounding},
\texttt{confounder\_guardrail}, \texttt{mechanism},
\texttt{ticket\_decision}, and \texttt{json\_contract}.  The anchor is the
visible validation response on held-out cases; the kernel proxy is residual
energy weighted toward candidates with weak visible anchor; and the
involutive-closure gap records section pairs absent from the accepted edit
basis.

In a seed-42 trace with seven candidate steps, five accepted and two rejected,
the flat bracket residual already separated rejected candidates:
\[
  \overline{R}_{\mathrm{flat}}(\mathrm{accepted})=0.9299,
  \qquad
  \overline{R}_{\mathrm{flat}}(\mathrm{rejected})=1.4848.
\]
The algebroid decomposition sharpened the interpretation:
\[
\begin{array}{lcc}
\toprule
 & \mathrm{accepted} & \mathrm{rejected}\\
\midrule
\text{visible anchor norm} & 0.5336 & 0.3641\\
\text{kernel proxy norm} & 0.5194 & 1.0377\\
\text{involutive-closure gap} & 0.0000 & 0.3333\\
\bottomrule
\end{array}
\]
The kernel proxy predicted rejection with AUC $0.8$, while the
involutive-closure gap separated the two rejected candidates in this small
trace.  The rejected latent section labels were
\[
  \texttt{confounder\_guardrail+evidence\_grounding}
  \quad\text{and}\quad
  \texttt{confounder\_guardrail+ticket\_decision}.
\]
These labels are semantically meaningful: they suggest that the problematic
edits were not arbitrary bad patches, but missing closure conditions linking
evidence scope, confounder handling, and admission/review decisions.  The next
test is to materialize these latent sections as explicit subskills and check
whether doing so reduces kernel energy and held-out transfer regressions.

A live follow-up \SkillOpt\ run on the same Odyssey/Democritus causal-claim
skill provides an additional sanity check on the validation-gate interpretation.
At epoch 2, step 5 of a 16-step run, the train minibatch contained one failure
and three successes, yielding hard accuracy $0.7500$ and soft score $0.7107$.
Reflection and aggregation produced three candidate edits, the selector kept
two, and the proposed patch expanded the skill from 2270 to 2850 characters.
On the seven-item held-out selection set, however, the candidate reached hard
accuracy $1.0000$ and soft score $0.9468$, but its mixed score
$0.9734$ remained below the incumbent $0.9836$, so the edit was rejected.  The
run therefore illustrates the desired discipline: even a patch that looks
perfect by hard held-out accuracy can be rejected when the softer validation
signal detects a regression.

\begin{center}
\begin{tabular}{lcccc}
\toprule
Step & Train hard & Train soft & Selection hard/soft & Action \\
\midrule
5/16 & 0.7500 & 0.7107 & 1.0000 / 0.9468 & reject: $0.9734 \leq 0.9836$ \\
6/16 & 1.0000 & 0.9807 & in progress & pending \\
\bottomrule
\end{tabular}
\end{center}

Step 6 is still incomplete in the available transcript.  Its train minibatch
had four successes, produced two success-derived edits, and proposed a shorter
patch from 2270 to 2522 characters.  The first three held-out evaluations were
mixed: one perfect item, followed by a hard failure on
\texttt{democritus\_wapo\_elnino} and a partial score on
\texttt{democritus\_wapo\_emperorpenguin}.  We therefore treat this as a
provisional trace rather than a final result.  If the completed run rejects
this patch as well, it would strengthen the algebroid diagnostic: high
minibatch performance and success-only reflection can still move hidden
workflow state in a way that the held-out anchor detects as path-dependent
regression.

\section{Conclusion}

Agentic skill optimization is not ordinary prompt tuning.  A skill edit acts at
an anchor in a typed artifact, propagates through a workflow, receives feedback
from rollout and validation, and may fail to commute with other seemingly local
repairs.  Lie algebroids give a compact language for this structure: controlled
edit policies live upstairs, visible artifact changes appear through the
anchor, and kernel directions record hidden procedural state that may matter
only in later composition.

 This paper introduced a new approach to skill optimization based on Lie algebroids. \LASKO\ models agentic
skill optimization as optimization over a Lie algebroid of controlled edit
policies above a tangent category of typed Markdown workflows.  Forward tangent
structure pushes edits through the workflow; reverse tangent structure pulls
critiques and validator residuals back to anchors; bracket and curvature
diagnostics measure order-dependence in both artifact changes and blame
assignment.  Stable skills are the low-curvature fixed points of this
forward-backward repair process. Preliminary results reported in this paper show order-of-magnitude improvements in performance, as cheap inexpensive Lie-bracket screening tests filter out a large percentage of edits before requiring expensive server-hosted LLM validations. 

\appendix

\section{Implementation Notes: Bracket Probes, Served Validation, and Audit}

The experiments in this paper distinguish two very different costs.  A
\emph{cheap bracket probe} is a static, application-specific proxy for whether
two controlled sections are likely to fail to commute.  It does not execute the
agent, call a served model, or validate a complete repair program.  It inspects
local structure such as read/write overlap, shared anchors, residual overlap,
context-chart overlap, witness overlap, or query-scope pressure.  A
\emph{served validation} is an expensive semantic check of an actual candidate
repair sequence, typically through an OpenAI-compatible endpoint hosted by
Exo.  The served validator receives the candidate sequence and an edge catalog,
then returns the ordered residual edges it judges closed.

This distinction is important for interpreting the probe counts.  For example,
in the WaPo coffee PSR-LASKO run, brute-force ordered-pair validation reaches
score \(1.000\) using \(73\) validation probes and no bracket probes.  The
PSR-LASKO policy reaches the same score using \(72\) cheap bracket probes and
only \(19\) served validation probes.  The bracket probes are not additional
model validations; they are the inexpensive geometric prefilter that decides
which candidate repair orders deserve served validation.

The wall-clock measurements make the distinction concrete.  On the served
WaPo coffee run, the \(72\) PSR-LASKO bracket probes took \(0.000508\) seconds
in total, or about \(7.1\) microseconds per probe.  The \(19\)
Exo-hosted validations took \(122.086\) seconds in total, with a mean served
latency of \(6.426\) seconds per validation.  At that measured latency,
validating all \(73\) ordered-pair candidates through the same served endpoint
would cost approximately \(469\) seconds before accounting for retries or
endpoint variability.  Thus the bracket layer is not competing with the
served validator; it is a near-zero-cost routing layer that decides where the
served validator is worth spending.

The Lovesac product-feedback PSR foundry repeats this separation on a
non-scientific review artifact.  The \(72\) bracket probes took \(0.000127\)
seconds, while the \(19\) served validations took \(212.396\) seconds.  The
brute-force reference needed \(73\) served validation probes for the same score
\(1.000\).  This keeps the empirical claim narrow: LASKO does not remove
semantic validation, but it changes the validation queue before the expensive
served calls are made.

We also ran a partial overnight model sweep on the same WaPo coffee task using
downloaded local Exo models.  The sweep used the per-model
\texttt{psr\_lasko\_results.json} files as the source of truth.  Across eight
evaluated served models, every model exhibits the same validation count
reduction: \(73\) exhaustive served validations versus \(19\) LASKO served
validations, with \(72\) cheap bracket probes.

\begin{center}
\begin{tabular}{lrrrr}
\toprule
Served model & Vanilla time & LASKO time & Score & Speedup\\
\midrule
DeepSeek V3.1 4-bit & 538.1 s & 36.2 s & 0.578 & \(14.85\times\)\\
Nemotron 70B 8-bit & 712.4 s & 86.0 s & 0.474 & \(8.28\times\)\\
Nemotron Nano 4-bit & 4075.7 s & 515.3 s & 0.161 & \(7.91\times\)\\
Llama 3.2 3B 4-bit & 1365.2 s & 242.7 s & 0.000 & \(5.63\times\)\\
Qwen3 0.6B 4-bit & 1927.4 s & 359.8 s & 0.670 & \(5.36\times\)\\
Qwen3 Next 80B 6-bit & 215.6 s & 38.4 s & 0.869 & \(5.61\times\)\\
Qwen3.5 2B 8-bit & 3283.8 s & 776.4 s & 0.000 & \(4.23\times\)\\
gpt-oss 20B MXFP4-Q8 & 1210.7 s & 328.0 s & 1.000 & \(3.69\times\)\\
\bottomrule
\end{tabular}
\end{center}

The observed speedups range from \(3.69\times\) to \(14.85\times\), with mean
speedup \(6.94\times\).  Across the evaluated lanes, the total served
validation time saved is \(10946.0\) seconds, while the largest measured
bracket-screening time is \(0.000909\) seconds.  The ICML tutorial claim is
therefore deliberately narrow but concrete: when the same repair task is
served across multiple local models, LASKO consistently replaces most expensive
served validations by sub-millisecond static bracket screening.
The score column should be read separately from this routing claim.  It is a
served-LLM semantic repair score, so it varies with model capacity, instruction
following, JSON/edge-ID format compliance, and stochastic endpoint behavior.
LASKO does not guarantee that every local model is an equally good semantic
repair validator; it guarantees that the same structural bracket screen can
route the expensive validation calls before those model-specific effects are
paid for.

\subsection{Application-specific bracket proxies}

The formal role of the bracket is generic:
\[
  [s_i,s_j] \approx s_i s_j - s_j s_i .
\]
Operationally, however, the proxy used to rank section pairs depends on the
artifact being optimized.  In PSR-LASKO, the proxy combines read/write
dependencies between PSR sections, overlap in residual components
\((R_H,R_{\mathrm{res}},R_W,R_{\mathrm{causal}},R_{\mathrm{scope}},
R_{\mathrm{DB}})\), context-chart interactions, witness dependencies, Hankel
row/column effects, and drift pressure induced by off-query clusters.  In
Markdown \SkillOpt, the analogous proxy may use shared anchors, schema fields,
examples touched by both edits, validator dependencies, and hidden tool or
prompt-routing state.  In \BASKET/\ROCKET, it may use dependencies from table
extraction to claims, claims to mechanism graphs, evidence ledgers to
citations, and year-over-year stability constraints.

Thus the invariant idea is not a single universal formula for the cheap probe.
It is the architecture:
\[
  \text{application-specific static noncommutation proxy}
  \quad\Longrightarrow\quad
  \text{focused expensive validation}.
\]

\subsection{Served validation protocol}

For the served experiments, candidate repair sequences are rendered as compact
JSON objects containing section names, anchors, read/write fields, residual
families, and an explicit catalog of allowed ordered edges.  The served model
is instructed to return a strict JSON object of the form
\[
  \{\texttt{"closed\_edges"}:[\cdots]\}.
\]
An edge is counted as closed only when both named sections appear in the
candidate sequence and the left section occurs earlier than the right section.
The model is not asked to invent edges, score the science, or infer additional
causal structure.  Its task is a deliberately narrow validation step over a
given catalog.

This design separates three layers:
\begin{enumerate}
  \item the deterministic oracle edge computation used for calibration;
  \item the cheap bracket screen used to choose candidate pairs;
  \item the served validator used to check selected sequences under model
  latency, parsing, and endpoint constraints.
\end{enumerate}
The served path therefore tests not only the mathematical prioritization, but
also whether a real hosted model can execute the validation contract reliably.

\subsection{WaPo coffee PSR-LASKO audit}

The WaPo coffee run exposed two practical issues that are useful for future
experiments.  First, a short token budget caused the served model to reason in
prose and truncate before emitting the requested JSON object.  The benchmark
was updated to use a larger maximum token budget and a stricter prompt stating
that the first character must be \texttt{\{}.

Second, the parser initially recovered edge identifiers too aggressively from
prose.  If a response listed both closed and not-closed edges, a naive fallback
could recover identifiers mentioned in negative statements.  The parser was
tightened so fallback recovery ignores lines containing ``not closed'' and
accepts only lines that explicitly mark an edge as closed or added.  The final
served WaPo coffee run produced \(48\) evaluation records: \(47\) strict-JSON
responses and one audited text fallback.  The resulting scores were:
\[
\begin{array}{lcccc}
\toprule
\text{Method} & \text{Bracket probes} & \text{Served validations} &
\text{Score} & \text{Faithfulness}\\
\midrule
\text{Greedy repair} & 0 & 10 & 0.000 & 1.000\\
\text{Random ordered pairs} & 0 & 19 & 0.371 & 1.000\\
\text{Brute-force reference} & 0 & 73 & 1.000 & \text{oracle}\\
\text{PSR-LASKO bracket screening} & 72 & 19 & 1.000 & 1.000\\
\bottomrule
\end{array}
\]
For the PSR-LASKO served row, those \(72\) bracket probes took \(0.000508\)
seconds, while the \(19\) served validations took \(122.086\) seconds.  The
random same-budget served row took \(125.156\) seconds for the same number of
served validations but reached score \(0.371\).  This is the operational
version of the geometric claim: cheap infinitesimal compatibility tests do
not replace validation, but they dramatically change which validations are
performed.
This audit is part of the result.  It shows that LASKO's claimed reduction is
not merely an offline scoring artifact: the same candidate sequence can be
checked through a real served model, with parse behavior and faithfulness
recorded for inspection.

\end{document}